\documentclass{article}
\usepackage[utf8]{inputenc}
\usepackage[top=1 in,bottom=1 in, left=0.88 in, right=0.88 in]{geometry}
\twocolumn  
\usepackage{multicol}

\usepackage{times}
\usepackage{soul}
\usepackage{url}
\usepackage{hyperref}
\usepackage{amsmath}
\usepackage{amsthm}
\usepackage{amsfonts}  
\usepackage{booktabs}
\usepackage{multirow}
\usepackage{algorithm}
\usepackage{algorithmic}
\usepackage[switch]{lineno}
\usepackage{graphicx}
\usepackage{caption}
\usepackage{subcaption}
\usepackage{amssymb}
\usepackage[section]{placeins}
\usepackage{float}
\usepackage{sectsty}

\sectionfont{\fontsize{14}{15}\selectfont}

\begin{document}

\twocolumn
\title{A Sequential Optimal Learning Approach \\
to Automated Prompt Engineering in Large Language Models}
\author{Shuyang Wang$^{1}$ \and Somayeh Moazeni$^{2}$ \and Diego Klabjan$^{3}$}
\date{\normalsize $^1$Department of Engineering Sciences and Applied Mathematics, 
Northwestern University\\
$^2$School of Business, Stevens Institute of Technology \\
$^3$Department of Industrial Engineering and Management Sciences, 
Northwestern University}

\maketitle

\begin{abstract}
Designing effective prompts is essential to guiding large language models (LLMs) toward desired responses. Automated prompt engineering aims to reduce reliance on manual effort by streamlining the design, refinement, and optimization of natural language prompts. This paper proposes an optimal learning framework for automated prompt engineering, designed to sequentially identify effective prompt features while efficiently allocating a limited evaluation budget. We introduce a feature-based method to express prompts, which significantly broadens the search space. Bayesian regression is employed to utilize correlations among similar prompts, accelerating the learning process. To efficiently explore the large space of prompt features for a high quality prompt, we adopt the forward-looking Knowledge-Gradient (KG) policy for sequential optimal learning. The KG policy is computed efficiently by solving mixed-integer second-order cone optimization problems, making it scalable and capable of accommodating prompts characterized only through constraints. We demonstrate that our method significantly outperforms a set of benchmark strategies assessed on instruction induction tasks. The results highlight the advantages of using the KG policy for prompt learning given a limited evaluation budget. Our framework provides a solution to deploying automated prompt engineering in a wider range applications where prompt evaluation is costly.
\end{abstract}


\section{Introduction}\label{sec:intro}
Large Language Models (LLMs) demonstrate exceptional capabilities in following instructions, making them a powerful tool to various downstream tasks \cite{sahoo2024systematic,liu2023pre,brown2020language,radford2019language}. A well-designed prompt steers an LLM to generate desired responses, enabling effective adaptation to downstream applications without incurring the high cost of fine-tuning the model weights. Nevertheless, creating effective prompts can be challenging due to the sensitivity of LLM outputs to prompt variations \cite{sclar2024quantifying,lu2022fantastically,zhao2021calibrate}. In addition, manually identifying ideal prompts is often time-consuming and lacks systematic guidance. Automated approaches to designing, optimizing, and refining LLM prompts mitigate this challenge by minimizing the need for manual intervention.

Recent efforts in automated prompt engineering primarily have focused on iterative evaluation and refinement in order to converge to ideal prompts \cite{guo2024connecting,zhou2023large,pryzant2023automatic,prasad2022grips,shin2020autoprompt}. The methods by these studies often assume the availability of numerous iterations. However, in many real-world scenarios, opportunities to evaluate prompts are limited, as each prompt evaluation is costly or time-consuming. For example, in medical research, each prompt evaluation could involve extensive time and resources from medical professionals, making it impractical to test a large number of variations before selecting the final one. 

Moreover, many existing approaches search over a set of precrafted candidate prompts for ideal prompts \cite{shi2024best,zhou2023large,luo2023prompt} These methods require identifying and enumerating a set of prompts, which restricts the scalability as the candidate set expands. Furthermore, it fails to utilize the correlation among similar prompt to expedite the learning.

To fully unlock the potential of LLMs across diverse scenarios, it is crucial to develop an automated prompting framework that is capable of capturing dependencies among prompts and efficiently identifying high-performing prompts within few evaluations. This paper presents a principled forward-looking iterative process for automated prompt engineering through the optimal design of a sequence of prompts.

We propose an interpretable feature-based approach to prompt representation. Various categorical or numerical features can be considered to characterize detailed aspects of a prompt, such as the selection and ordering of demonstrative examples. Previous works identify factors within a prompt that influence the LLM outputs but often treat these aspects in isolation. We allow for capturing various interactions among prompt attributes and can accommodate potential constraints on the features. This feature-based prompt representation enables the inclusion and exploration of a vast and diverse set of prompts, which, unlike previous approaches, do not require manual prompt provision. In addition, in contrast to prompt descriptions based on embedding vectors, our representation is inherently interpretable.

We adopt a Bayesian approach to refine beliefs about the influence of prompt features on the LLM response. This approach supports the integration of prior knowledge and user opinions as well as enabling to capture feature correlations. To operationalize this, we define a probabilistic model to link prompt features to a response quality of interest. In this paper, we demonstrate our approach using LLM response accuracy as the primary quality metric.

Next, we formalize the iterative process of automated prompt engineering in the presence of limitations on the number of prompt evaluations as a sequential decision-making problem. Given the limited opportunities for prompt evaluation, this problem falls into the category of finite-horizon discrete-time Markov decision processes; see \cite{puterman2014markov}. An optimal learning policy sequentially selects a feasible prompt representation for evaluation, aiming to maximize the expected outcome of the final prompt. Due to the potentially large prompt feature space, the curse of dimensionality \cite{powell2022reinforcement} hinders the exact computation of the optimal prompt selection. We adopt an approximate policy for the optimal learning problems, known as the expected improvement policy in \cite{chick2006bayesian,chick2010sequential} or Knowledge-Gradient (KG) policy in \cite{frazier2009knowledge,powell2012optimal}. This is a forward-looking policy that maximizes the expected improvement in the value of information in each learning phase. The KG policy often excels in practical scenarios
with limited evaluation budgets, frequently outperforming common static data acquisition strategies and dynamic test-and-learn policies. For the consistency of the KG policy, see \cite{frazier2011consistency} for correlated features and \cite{han2016optimal} for constrained search space.

The large space of prompt candidates, defined by constrained features, makes it impractical to enumerate all feasible alternatives for determining KG decisions. To address this challenge, we leverage recent advancements in scalable optimal learning and KG computation. In contrast to earlier results to compute KG decisions \cite{frazier2009knowledge} based on enumeration of all feasible alternatives, recent computational methods \cite{moazeni2020sequential,han2016optimal,defourny2015optimal} build on optimal quantization of the response probability followed by mixed-integer conic optimization reformulations to leverage efficient optimization solution methods. For optimal learning problems with larger feature spaces, an iterative process involving solving mixed-integer linear optimization problems is employed to achieve even greater computational efficiency and scalability.

Our framework allows for different prompt representations and selection policies, hence encompassing various existing methods for automated prompt engineering. For example, when a small, finite set of precrafted prompt templates is provided and a point-wise utility model is used, our setting simplifies to the setup in \cite{luo2023prompt,zhou2023large,shi2024best}. 

We assess the performance of sequential prompt learning with adaptive prompt selection policies on a dataset of instruction induction tasks \cite{honovich2022instruction}. This benchmark dataset contains 24 instances of instruction induction, designed for LLMs to deduce implicit tasks or instructions from language prompts including answers or contextual information. For the instruction induction tasks, we first propose a feature-based prompt template and use the accuracy of responses collected from GPT-3.5 on the validation data as the primary performance metric. For the prompt selection, we evaluate the KG policy, the adaptive myopic policy, the increasingly popular Thompson sampling policy, and a number of other automated prompt engineering methods such as EvoPrompt \cite{guo2024connecting} using an evolutionary algorithm to refine prompts and TRIPLE \cite{shi2024best} using a multi-armed bandit approach to select prompts.

Our analyses show the effectiveness of our approach particularly with the KG prompt selection policy, which is capable to converge to high-quality prompts within 30 or fewer prompt evaluations. These prompts significantly outperform those generated by the benchmarks on the test data using the same number of LLM interactions. Further analysis reveals that the KG policy is particularly favorable for challenging tasks with high uncertainty in LLM responses and significant sensitivity to prompts, achieving a substantial margin over baseline policies. Our findings highlight the advantages of using the KG policy in prompt engineering to selectively evaluate prompts, expanding the potential for deploying automated prompt engineering in applications with large prompt evaluation costs.

Our contributions can be summarized as follows.
\begin{itemize}
    \item We introduce a sequential optimal learning framework for automated prompt engineering to guide through the process of designing a sequence of prompts that effectively elicit accurate responses from an LLM. The approach is particularly effective for applications where prompt evaluation is resource-intensive.
    \item We propose a feature-based approach to represent language prompts, which greatly expands the prompt search space. A link function maps the features to LLM response accuracy through Bayesian model parameters to leverage correlations among prompts with shared characteristics. Our method enables simultaneous optimization of multiple features to generate improved prompts.
    \item We leverage the KG policy within our sequential prompt learning to efficiently identify high-performing prompts in large prompt spaces. The KG policy outperforms various benchmark policies, especially for challenging tasks with high uncertainty in LLM response.
\end{itemize}

The remainder of the paper is organized as follows: Section~\ref{sec:relatedwork} reviews the related literature. Section~\ref{sec:problem} discusses the generic iterative process for automated prompt learning, formalizing the problem as a sequential decision making problem. Section~\ref{sec:optimallearning} discusses the forward-looking KG policy for the prompt selection in iterative automated prompting. Illustrative examples and computational results are provided in Sections~\ref{sec:experiments} and \ref{sec:analysis}. Finally, conclusions and potential extensions are discussed in Section~\ref{sec:conclusion}.
\section{Related Work}\label{sec:relatedwork} 

\paragraph{Generation-then-selection Methods}
\cite{zhou2023large} present a two-phase pipeline of instruction generation and selection. This method generates a set of candidate instructions, which are evaluated and filtered based on their performance on the downstream tasks until the best one from the candidate set is found. \cite{luo2023prompt} uses an optimal control paradigm to systematize the process of iteratively updating and selecting from a set of candidate prompts. However, these approaches are limited to search spaces represented by individual prompt candidates. Our proposed framework encompasses different prompt representations and exploration policies. Our feature-based approach to represent prompts captures their dependencies and enables a diverse search space. \cite{shi2024best} follows the two-phase pipeline with a focus on prompt selection subject to a fixed evaluation budget. They formulate the selection as a multi-armed bandit (MAB) problem and utilize the continuously reject method to select from the candidate set. However, their solution is not scalable with the size of the prompt candidate set. In contrast, our framework with the KG prompt selection policy can accommodate larger spaces of prompts.

\paragraph{Edit-based Methods}
An approach to automated prompting focuses on generating refined prompts by iteratively editing base prompts. \cite{prasad2022grips} propose GrIPS, which iteratively applies text-based edit operations, such as word substitutions and deletions, to a base prompt. The best candidate is then selected as the new base prompt. Alternatively, \cite{guo2024connecting} apply evolutionary algorithms and generate new candidate prompts by performing mutation and crossover operations using an LLM, retaining high-quality prompts for the next generation. \cite{fernando2024promptbreeder} also use an LLM to perform mutation operations on a population of prompts but employ a self-referential way of improving both the prompts and the mutation operations. \cite{jin2024apeer} propose an iterative prompt refinement scheme specially crafted for relevance ranking in information retrieval. While these attempts show the potential of edit-based approaches for generating high-quality prompts, they mainly rely on local search by modifying existing prompts. Our method, however, explores the prompt space in a forward-looking principled manner using Bayesian optimal learning, and utilizes knowledge from prior observations to inform future selections of prompts to evaluate.

\paragraph{Prompt Learning Methods} Recent works \cite{chen2024instructzero,lin2024use} experiment using Bayesian Optimization to search in the embedding space of prompts, but white-box LLMs are required to facilitate the optimization steps. Our approach also builds on Bayesian learning, but we directly search in the space of discrete prompts and eliminate the need for white-box LLMs. The concept of prompt learning is also used in \cite{mingkai2022rlprompt}, which trains a reinforcement learning model to select tokens as actions to form prompts. The training stage requires sufficient evaluation budget, while our framework is capable of learning from only a limited number of prompt evaluation.

\paragraph{Gradient-based Methods}
Gradient-based algorithms have been used to solve the problem of optimizing the prompt performance over the prompt space. \cite{shin2020autoprompt,shi2023toward} model prompts as sequences of trigger tokens, and compute the gradients of the log-likelihood of the language model generating the target outputs with respect to the embeddings of candidate tokens to guide the search for optimal tokens. \cite{wen2024hard} compute the gradients of a similarity metric between the generated and target outputs with respect to the prompt embeddings to guide prompt optimization, and project the optimized embeddings to discrete prompts using nearest neighbors. These methods require access to the internal parameters of the LLM, making them incompatible with black-box LLMs. Moreover, these methods require computationally intensive gradient computations. In contrast, our approach generates human-readable prompts without accessing internal parameters.
\section{Sequential Optimal Prompt Learning}\label{sec:problem}
\begin{figure*}[tb]
\centering   
\includegraphics[width=0.8\linewidth]{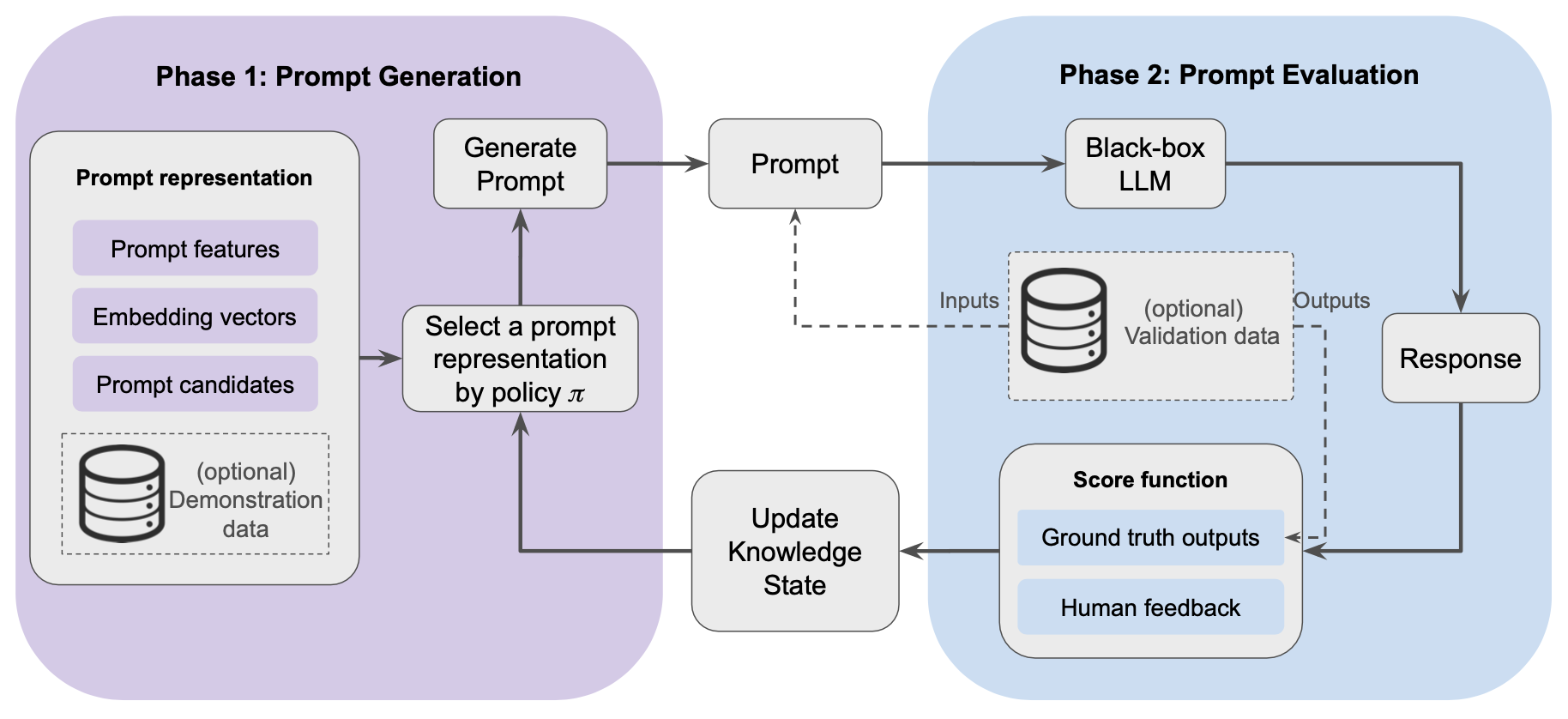}
    \caption{SOPL: Sequential optimal prompt learning for automated prompt engineering}
    \label{fig:diagram}
\end{figure*}

We introduce a sequential optimal prompt learning framework, called \emph{SOPL}, for automated prompt engineering that is compatible with black-box LLMs and generates human-readable prompts, while addressing the challenge of limited number of iterations for prompt evaluation. Our framework, depicted in Figure~\ref{fig:diagram}, follows the iterative process outlined in Algorithm~\ref{algo:bayesianframework}. The components of the framework are explained in the subsequent subsections. 

\begin{algorithm}[tb]
\caption{Sequential optimal prompt learning}
\label{algo:bayesianframework}
\textbf{Require} Maximum iteration $N$, score function $\mathrm{Eval}: \mathcal{X} \to (0, 1)$, prompt representation selection policy $\pi: \mathcal{S} \to \mathcal{X}$, Bayesian update function $\operatorname{Update}: \mathcal{S} \times \mathbb{R} \to \mathcal{S}$ that implements \eqref{eq:updateequations-begin}-\eqref{eq:updateequations-end}.
\begin{algorithmic}[1]
\STATE Initialize knowledge state $S$.
\STATE Initialize best prompt so far $x^*$.
\STATE Initialize best score so far $u^* \gets -1$.
\FOR{step $n = 1, ..., N$}
\STATE Get prompt representation $x \gets \pi(S)$.
\STATE Evaluate prompt and get a score $u_x \gets \mathrm{Eval}(x)$.
\IF{$u > u^*$}
\STATE Record best score so far $u^* \gets u_x$.
\STATE Record best prompt so far $x^* \gets x$.
\ENDIF
\STATE Update knowledge state $S \gets \mathrm{Update}(S, \operatorname{logit}(u_x))$.
\ENDFOR
\STATE \textbf{return} $x^*$.
\end{algorithmic}
\end{algorithm}

\subsection{Feature-Based Prompt Representation}\label{sec:feature-based}
To identify high-quality prompts, it is essential to explore a diverse set of prompts tailored to the specific downstream task. We adopt a feature-based representation for prompts, expressing a language meta prompt through various features that capture its content and structure. Prompt features are denoted by the vector $x$ that specifies a textual prompt. Examples of such features include the structural template, tone, role, context, demonstrations, specificity, complexity, embeddings, task type, question framing, constraints, and temporal references. These features are generally either manually engineered by the user or derived from established prompt templates in the literature corresponding to the task. Refer to Section 5.1 for the specific features and categories utilized in the instruction induction task for our experiments, developed based on the template proposed in \cite{honovich2022instruction}. 

Previous studies have primarily examined each feature in isolation, whereas we integrate these features in the prompt representation to leverage the potential synergies that emerge from their combination. By enriching the meta prompt with multiple features known to influence LLM responses, we can expand the search space.

In general, the feature space may contain variables of different types: continuous, categorical, and ordinal. In addition, various requirements may  be imposed on the features either by definition or by the user's preferences to account for mutually exclusive features, conditional features, combined effects of multiple features, disjunctive features, and multiple-choice decisions. The set of feasible feature combinations forms a diverse and potentially large search space, denoted by $\mathcal{X}$. Our framework does not require explicitly identifying and enumerating all feasible prompts. Instead, the feasible prompt space $\mathcal{X}$,  which encompasses the feasible values of the prompt features, is specified solely by linear inequality or equality constraints.

Our framework encompasses existing approaches as special cases. For example, the methods \cite{zhou2023large,shi2024best,luo2023prompt} that follow the pipeline of generating and selecting from a set of candidates can be thought of as using the candidate set as $\mathcal{X}$.

\subsection{Prompt Evaluation}
The prompt evaluation phase involves querying a black-box LLM with the prompt constructed from $x$ and collecting the LLM's response. When evaluating the prompt on the downstream task, we measure the quality of the observed LLM response to the prompt associated with $x$ using a numeric score $u_x$, which is computed from the score function $\operatorname{Eval}$ defined on $\mathcal{X}$; see Algorithm 2 for details on $\operatorname{Eval(x)}$. Given labeled data, the score can be computed by comparing the LLM responses to the ground truth labels and calculating the percentage of accurate response. For downstream tasks that require human evaluation, the score is based on human feedback. 

For common metrics such as accuracy, F1-score, and point-wise mutual information \cite{bouma2009normalized}, the score lies in the interval between 0 and 1. We assume that \begin{align}
    \eta_x := \operatorname{logit}(u_x)  = \Theta^\top x + \epsilon, \label{eq:utilitymodel}
\end{align}
where $\Theta \sim \mathcal{N}(
    \mu_{\Theta}, \Sigma_{\Theta}/\rho)$ is the $D$-dimensional model parameter and $\epsilon \sim \mathcal{N}(0, 1/\rho)$ with variance $1/\rho$ is the measurement noise. The quantity $\eta_x$ represents the utility of $x$. The mean $\mu_{\Theta}$ and the precision $\rho$ are the unknown parameters to estimate. For general score functions that return values beyond the interval between 0 and 1, alternative link functions can be employed in \eqref{eq:utilitymodel} in place of the logit function.

\subsection{Knowledge State Update}
An approach to addressing uncertainty in the effectiveness of prompt features is Bayesian learning. The Bayesian approach to inference can account for multiple levels of randomness and correlation by using prior distributions for model parameters. Additionally, existing knowledge and user input can be incorporated into these prior probability distributions. We adopt the Bayesian framework, letting a multivariate normal prior for the coefficients, $\mu_{\Theta}$, and a Gamma prior for the precision, $\rho$, i.e.,

\begin{align}
    \mu_{\Theta} | \rho &\sim \mathcal{N}\left(\theta, \Sigma/\rho\right) \label{eq:thetadistribution} \\
    \rho &\sim \text{Gamma}(a, b).
\end{align}

The belief represented by $S = (\theta, \sigma, a, b)$ is referred to as the \emph{knowledge state}. The knowledge state encodes prior observations of prompt performance and serves as the basis to inform future decisions. The multivariate distribution \eqref{eq:thetadistribution} captures dependencies among unknown parameters, implying that learning about one prompt can provide insights into the effectiveness of several other prompts, thereby enhancing the learning speed. 

We iteratively update the knowledge state based on observed responses to queried prompts. At iteration $n$, the knowledge state is denoted by $S_n = (\theta_n, \Sigma_n, a_n, b_n)$, and the selected prompt representation is denoted by $x_n$. After the $n$-th iteration of prompt evaluation, we observe the score $u_n := u_{x_n}$ and update the knowledge state as follows.
\begin{align}
    \theta_{n+1} &= \theta_{n} + \frac{\operatorname{logit}(u_{n}) - \theta_{n}^\top x_{n}}{(1 + x_t^\top \left(\Sigma_{n} + \Sigma_{\Theta}\right) x_{n})}\Sigma_{n}x_{n} \label{eq:updateequations-begin}\\
   \Sigma_{n+1} &= \Sigma_{n} - \frac{\Sigma_{n}x_{n} x_{n}^\top \Sigma_{n}}{1 + x_{n}^\top\left(\Sigma_{n} + \Sigma_{\Theta}\right)x_{n}} \\
a_{n+1} &= a_{n} + \frac{1}{2} \\
b_{n+1} &= b_{n} + \frac{(\operatorname{logit}(u_{n}) - \theta_{n}^\top x_{n})^2}{2(1 + x_{n}^\top \left(\Sigma_{n} + \Sigma_{\Theta}\right) x_{n})} \label{eq:updateequations-end}
\end{align}

Equations~\eqref{eq:updateequations-begin}-\eqref{eq:updateequations-end} collectively define the mapping $\operatorname{Update}(S, \operatorname{logit}(x))$ appeared in line 11 of Algorithm~\ref{algo:bayesianframework}.

\subsection{Prompt Representation Selection Policy}
For each iteration, a prompt representation $x_n$ is selected according to a policy $\pi$. The goal is to find a prompt that maximizes the score on the downstream task after $N$ iterations of prompt evaluation.

Our framework allows for different prompt representation selection policies to explore the feasible prompt space $\mathcal{X}$ and update the knowledge state. Heuristic policies, such as adaptive myopic (Greedy) and Thompson sampling (TS), can be adopted. The Greedy policy selects the best prompt representatio $x$ based on the current knowledge state by \begin{equation}
    \pi^{Greedy}(S_n) := \underset{x \in \mathcal{X}}{\operatorname{argmax}}~\mathbb{E}[\eta_x | S_n] = \underset{x \in \mathcal{X}}{\operatorname{argmax}}~\theta_n^\top x\label{eq:greedy}.
\end{equation} The TS policy samples from the posteriors of the parameters by \begin{equation}
    \hat{\rho} \sim \operatorname{Gamma}(a_n, b_n),~~\hat{\theta} \sim \mathcal{N}\left(\theta_n, \Sigma_n/\hat{\rho}\right),
\end{equation} and then selects the prompt representation $x$ by \begin{equation}
    \pi^{TS}(S_n) := \underset{x\in\mathcal{X}}{\operatorname{argmax}} ~\hat{\theta}^\top x\label{eq:TS}.
\end{equation}
The heuristic policies such as Greedy and TS policies are adaptive, but they are not forward-looking, in the sense that they do not explicitly take into account the effect of selected prompts on the subsequent prompt selections and the overall learning process.

The prompt representation selection policy $\pi$ is a key building block influencing the performance of the SOPL framework. When the number of iterations is limited to $N$, the process of sequential optimal prompt learning can be formulated as a finite-horizon Markov decision process, where the action space $\mathcal{X}$ consists of prompt representations, and the state space $\mathcal{S}$ consists of knowledge states. In the next section, we discuss an approximate policy for the optimal prompt representation selection policy.

\section{KG Prompt Selection Policy}\label{sec:optimallearning}
We consider a forward-looking optimal learning policy designed to maximize the expected improvement in an approximated value of information during each iteration. This approach, known as the Knowledge-Gradient (KG) policy, offers an approximate solution to the MDP for prompt selection. For additional discussion and analysis, refer to \cite{gupta1996bayesian,chick2010sequential,frazier2008knowledge,powell2012optimal}. The value of information is measured by the expected single-period improvement, i.e., the difference between the values of the knowledge states $S_{n+1}$ and $S_n$ if the prompt representation $x_{n+1} = x$ is selected. Hence, at iteration $n$, the following KG quantity is maximized:
\begin{equation}
    \nu^n_{x} := \mathbb{E}\left[ V_N(S_{n+1})| S_n, x \right] - V_N(S_n). \label{eq:KGquantity}
\end{equation}
where $V_{N}(S)$ is the value of the optimal policy at iteration $N$ for any knowledge state $S$, i.e., $V_{N}(S)=\max_{\pi \in \Pi}V_{N}^{\pi}(S)=\max_{x\in\mathcal{X}}\mathbb{E}[\eta_x|S]$, where $\eta_x$ is as in equation (1). Recall that $S_{n+1}$ is the transition from state $S_n$ induced by the updating procedure in equations (4)-(7). The quantity $ \nu_x^n$ is the marginal value of one more prompt representation $x$ being queried. Its value is always nonnegative.

The decision of the KG policy selects the prompt representation that maximizes the KG quantity in equation \eqref{eq:KGquantity}:
\begin{equation}
    \pi^{KG}(S_n) := \underset{x\in\mathcal{X}}{\operatorname{argmax}}~\nu^n_x.
\end{equation}
For discussion on the related concepts of asymptotic optimality and statistical consistency of the KG policy, the reader is referred to \cite{frazier2008knowledge,frazier2011consistency} when $\mathcal{X}$ is specified in the enumerative form, and see \cite{han2016optimal} when $\mathcal{X}$ is represented in a constraint-based form.

For any $x\in\mathcal{X}$, the KG quanitity corresponding to model (1) is given by:
\begin{align}
\label{eq:nux}&&\nu_{x}^{n}=\mathbb{E}[\max_{y\in\mathcal{X}}(p_{y}^{n}+q_{y}^{n}(x)T_{2a_n})|S_n]-\max_{y\in\mathcal{X}}p_{y}^n
\end{align}
where $T_{2a_n}$ follows a student's t distribution with $2a_{n}$ degrees of freedom, and 
\begin{eqnarray}
&&p_{y}^{n}=\theta^{\top}y\\
&&q_{y}^{n}(x)=\sqrt{\frac{b_n}{a_n (1+x^{\top}(\Sigma_{n}+\Sigma_{\Theta})x)}}x^{\top}\Sigma_{n}y
\end{eqnarray}
The expectation in equation \eqref{eq:nux} is with respect to the one-dimensional random variable $T_{2a_n}$. 

The first term in the KG quantity in equation \eqref{eq:nux} can be approximated by 
\begin{eqnarray}
&&\sum_{j=1}^{J}w_j\max_{y\in\mathcal{X}}(p_{y}^{n}+q_{y}^{n}(x)t_j )
\end{eqnarray}
where $t_1,...,t_{J}\in\mathbb{R}$ is the sequence of points that minimizes the quadratic quantization error of the Voronoi quantizer for $T_{2a_n}$, $t_0=-\infty$, and $t_{J+1}=\infty$. The weights are defined as $w_j=F_{T_{2a_n}}\left(\frac{t_j+t_{j+1}}{2}\right)- F_{T_{2a_n}}\left(\frac{t_{j-1}+t_{j}}{2}\right)$ for $j=1,\cdots, J$. Here, $F_{T_{2a_n}}$ is the cumulative distribution function of $T_{2a_n}$. 
Therefore, the selected prompt representation based on the KG policy at state $S_n$ is computed by solving the following mixed-integer optimization problem:
\begin{align}
\max_{(\hat{x},\tau)\in\mathcal{X}^{+},\tau\leq M, x,y^1,...,y^{J}\in \mathcal{X}} \sum_{j=1}^{J}w_j \theta_{n}^{\top}y^j +\tau\\
\text{s.t.~} \|P_{n}^{1/2}\hat{x}\|_{2}\leq \sum_{j=1}^{J}w_j t_jx^{\top}\Sigma_{n} y^{j}\\
\tau\cdot 1_{m}-M(1_m -x)\leq \hat{x}\leq Mx
\end{align}
Here, $m$ is the dimensionality of the prompt representation features, and $P_{n}:=\frac{a_n}{b_n}(\frac{A^{\top}A}{h^{\top}h}+\Sigma_n+\Sigma_{\Theta})$, where $A$ and $h$ form the equality constraints of the feasible set $\mathcal{X}=\{x|Ax=h, Bx\leq g\}$. In this problem, $\mathcal{X}^{+}$, consisting of elements $(x,\tau)\in \mathbb{R}^{m}$, represents the homogenized version of the set $\mathcal{X}$. In the last constraint, $M$ denotes a large constant. 
For further details and a computationally efficient iterative algorithm only involving solving mixed-integer programming problems to solve this problem, see Propositions 6 and 7 in \cite{moazeni2020sequential}. 
\section{Computational Experiments}\label{sec:experiments}

We demonstrate the performance of the proposed SOPL framework on the instruction induction tasks \cite{honovich2022instruction}. The dataset consists of 24 individual tasks, covering various aspects of text comprehension. Each data point comprises a pair of input and output. For example, for the task \texttt{larger\_animal}, one data point consists of an input ``cougar, flea" and an output ``cougar". The objective is to find an instruction such that when the LLM is queried with the instruction and an input, its response matches the correct output. A possible instruction for this task can be ``choose the larger animal". Similar to \cite{zhou2023large}, we generate possible instructions by prompting an LLM using a meta prompt, which consists of demonstrative examples of input-output pairs and asks for a possible instruction. For each task, we partition the dataset into three sets: a demonstration dataset, a validation dataset, and a held-out test dataset. 

\paragraph{Feature-Based Prompt Representation.}

We focus on five aspects of prompts that have been shown to impact LLM responses. \cite{zhou2023large} note that the template of the meta prompt impacts the effectiveness of the induced prompts; \cite{lu2022fantastically,zhao2021calibrate} find that both the selection and the order of demonstration examples influence generated texts; \cite{wu2023large,kong2024better} show that specifying different roles elicits diverse text generations from LLMs; \cite{deng2023rephrase,zhou2023large} discover that prompts paraphrased by LLMs yield improved performance on downstream tasks;  \cite{li2023large} observe that LLMs respond differently to different descriptions of tones.

We create a set of choices for each feature, summarized in Table~\ref{tab:features-details}. A feature vector $x$ specifies one choice for each feature. By applying one-hot encoding to represent each categorical feature, the prompt representation feature $x$ becomes a binary vector. All combinations of the prompt features, subject to the constraint that exactly one choice is selected for each feature, form the search space $\mathcal{X}$.

\begin{table*}
    \centering
    \begin{tabular}{c l }
    \toprule
        \textbf{Feature} & \textbf{Choices} \\
        \midrule
        Meta prompt template & 4 meta prompt templates shown in Figure~\ref{fig:meta_prompt}.\\
        \hline
        Demonstrative examples & 20 choices, each consists of 5 examples sampled from the demonstration dataset.\\
        \hline
        Roles & (Scientist, research assistant), (Professor, PhD student), (Mom, kid), \\
        &  (Programmer, AI system),  (Manager, employee), (I, friend), \\
        &  (Director, actor), (Coach, athlete), (Chef, sous chef). \\
        \hline
        Paraphrasing & Binary: if paraphrasing is used, we prompt the LLM again to generate \\
        & a variant of the instruction using the template in Figure~\ref{fig:paraphrase-template}. \\
        \hline
        Description & (empty), clear, detailed, simple, complex, precise, ambiguous, \\
        & technical, expository, conceptual, authoritative, friendly, formal, \\
        & informal, encouraging, stern, rude, assertive, humorous.\\
        \bottomrule
    \end{tabular}
    \caption{Prompt features used for instruction induction tasks}
    \label{tab:features-details}
\end{table*}

\begin{figure*}[tb]
    \centering
     \begin{subfigure}[b]{0.33\linewidth}
         \centering
         \includegraphics[width=\textwidth]{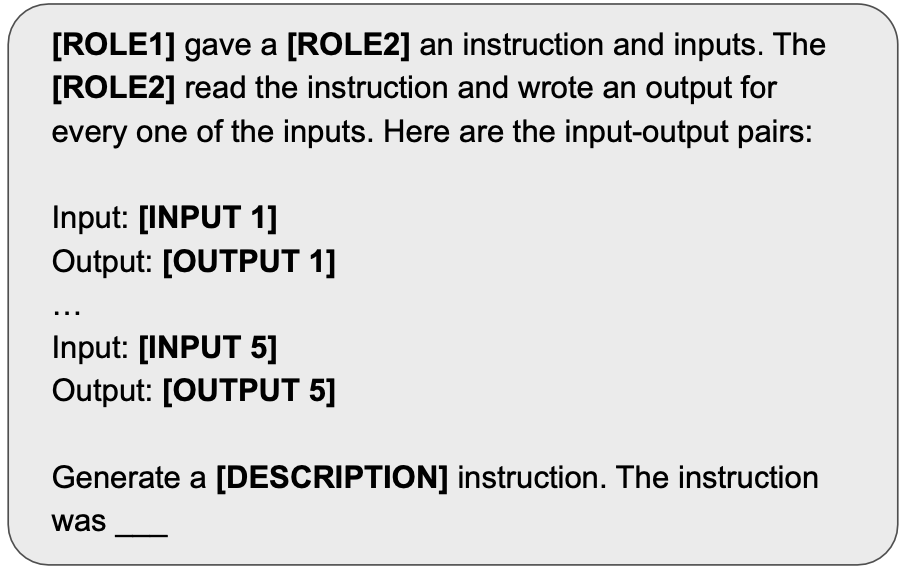}
         \caption{Meta Prompt Template 1}
     \end{subfigure}
     \hspace{0.1cm}
     \begin{subfigure}[b]{0.33\linewidth}
         \centering
         \includegraphics[width=\textwidth]{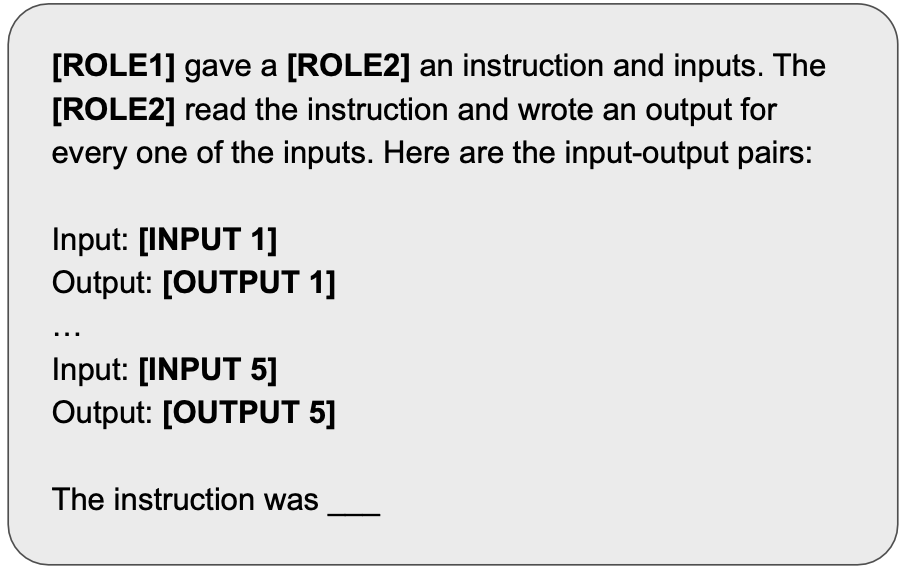}
         \caption{Meta Prompt Template 2}
     \end{subfigure}
     \par
     \centering
     \begin{subfigure}[b]{0.33\linewidth}
         \centering
         \includegraphics[width=\textwidth]{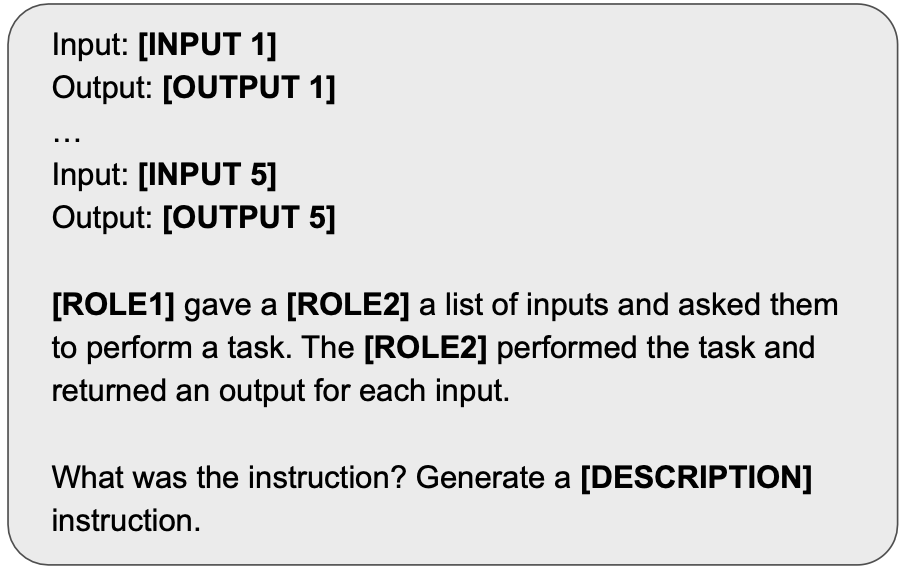}
         \caption{Meta Prompt Template 3}
     \end{subfigure}
     \hspace{0.1cm}
     \begin{subfigure}[b]{0.33\linewidth}
         \centering
         \includegraphics[width=\textwidth]{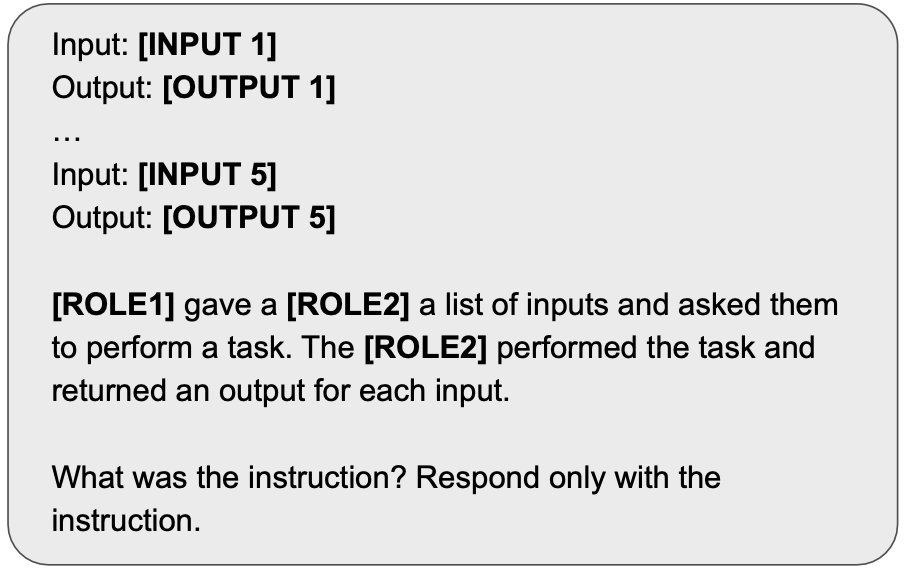}
         \caption{Meta Prompt Template 4}
     \end{subfigure}
     \caption{Meta Prompt Templates}
     \label{fig:meta_prompt}
\end{figure*}

\begin{figure}
    \centering
    \includegraphics[width=0.6\linewidth]{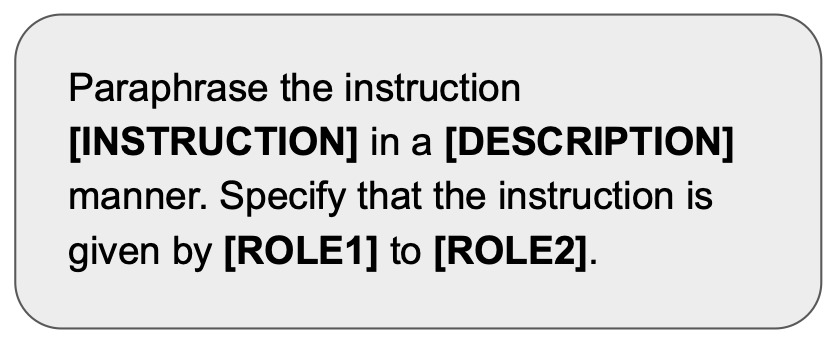}
    \caption{Paraphrasing Template}
    \label{fig:paraphrase-template}
\end{figure}

\begin{figure}
    \centering
    \includegraphics[width=0.6\linewidth]{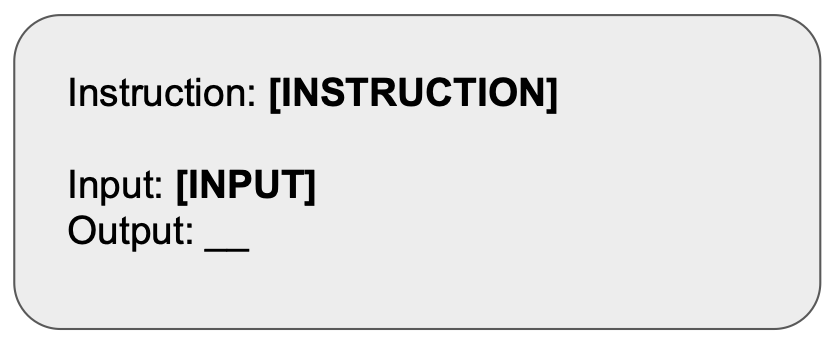}
    \caption{Evaluation Template}
    \label{fig:eval_prompt}
\end{figure}

\paragraph{Prompt Evaluation.} We evaluate the selected feature vector $x$ on the validation data and obtain the validation score $u_x$ by Algorithm~\ref{algo:evaluation}. We first convert the feature vector $x$ to a textual instruction in line 1 and 2. For example, if the feature vector specifies meta prompt template 1, 5 demonstrative examples, roles of I and friend, no paraphrasing, and description of clear, then we create a meta prompt by $\operatorname{MetaPrompt}(x)$, which inserts the 5 demonstrative examples in meta prompt template 1 in Figure~\ref{fig:meta_prompt}, and replaces [ROLE1] by ``I", [ROLE2] by ``friend", and [DESCRIPTION] by ``clear". We query an LLM with the meta prompt to generate an instruction $I_x$ that reflects the selected features. For each input $p_i$ in the validation data, we create an evaluation prompt by $\operatorname{EvalPrompt}(I_x, p_i)$, which combines the instruction and the input using the template in Figure~\ref{fig:eval_prompt}. The prompt is then used to query the LLM. A task-specific metric, such as exact match or F1-score defined in \cite{honovich2022instruction}, is used to compute a score by comparing the LLM response with the correct output $q_i$. The average score across all validation examples is used as the score $u_x$.

\begin{algorithm}[tb]
\caption{Score function $\operatorname{Eval} (x)$}\label{algo:evaluation}
\textbf{Require} An $\operatorname{LLM}$, validation data $\{(p_i, q_i)\}_{i=1}^{V}$, meta prompt construction function $\operatorname{MetaPrompt}$, evaluation prompt construction function $\operatorname{EvalPrompt}$, metric function $\operatorname{Metric}$ to evaluate LLM response.
\begin{algorithmic}[1]
\STATE Create meta prompt $M_x \gets \operatorname{MetaPrompt}(x)$.
\STATE Generate instruction $I_x \gets \operatorname{LLM}(M_x)$.
\FOR{$i = 1, ..., V$}
\STATE Get evaluation prompt $E_i \gets \operatorname{EvalPrompt}(I_x, p_i)$.
\STATE Receive LLM response $R_i \gets \operatorname{LLM}(E_i)$.
\STATE Evaluate LLM response $U_i \gets \operatorname{Metric}(R_i, q_i)$.
\ENDFOR 
\STATE Compute the average score $u_x = \frac{1}{V}\sum_{i=1}^{V} U_i$.
\STATE \textbf{return} $u_x$.
\end{algorithmic}
\end{algorithm}

\subsection{Benchmark Methods}
We compare our method with two benchmarks EvoPrompt \cite{guo2024connecting} and TRIPLE \cite{shi2024best}. Both methods provide solutions compatible with black-box access to LLMs and generate human-readable prompts. EvoPrompt uses the differential evolution algorithm to iteratively refine a population of prompts. TRIPLE employs the continuously reject algorithm to identify an effective prompt from a candidate pool under a fixed budget. In addition, we use Greedy and TS presented in \eqref{eq:greedy} and \eqref{eq:TS} as the baseline policies. 

\subsection{Implementation Details}
We record the instruction with the highest validation score during the process. When the maximum number of prompt evaluation is reached, the instruction with the highest validation score is used as the final instruction. The test score is obtained by evaluating the final instruction on the held-out test data, using a similar procedure in Algorithm~\ref{algo:evaluation}. We report the average test score across 20 replications with different random seeds.

The held-out test dataset for each task consists of 100 examples unless specified otherwise in \cite{honovich2022instruction}, and is kept the same for all replications. For each replication, we randomly select 10 examples from the rest of the data as the demonstration dataset, and then randomly select 100 examples or all remaining examples if fewer are available as the validation dataset.

We use OpenAI GPT-3.5 as the LLM for both generating and evaluating instructions. We allow $N=30$ opportunities to evaluate on the entire validation dataset. We set the population size to be $10$ for EvoPrompt as in \cite{guo2024connecting}, and set the size of the candidate pool to be $30$ for TRIPLE as in \cite{shi2024best}. We ensure that the same number of API calls to the LLM is used for evaluating on the validation data across all methods.

\section{Results and Sensitivity Analysis}\label{sec:analysis}

\begin{figure*}[tb]
    \centering
    \includegraphics[width=0.9\linewidth]{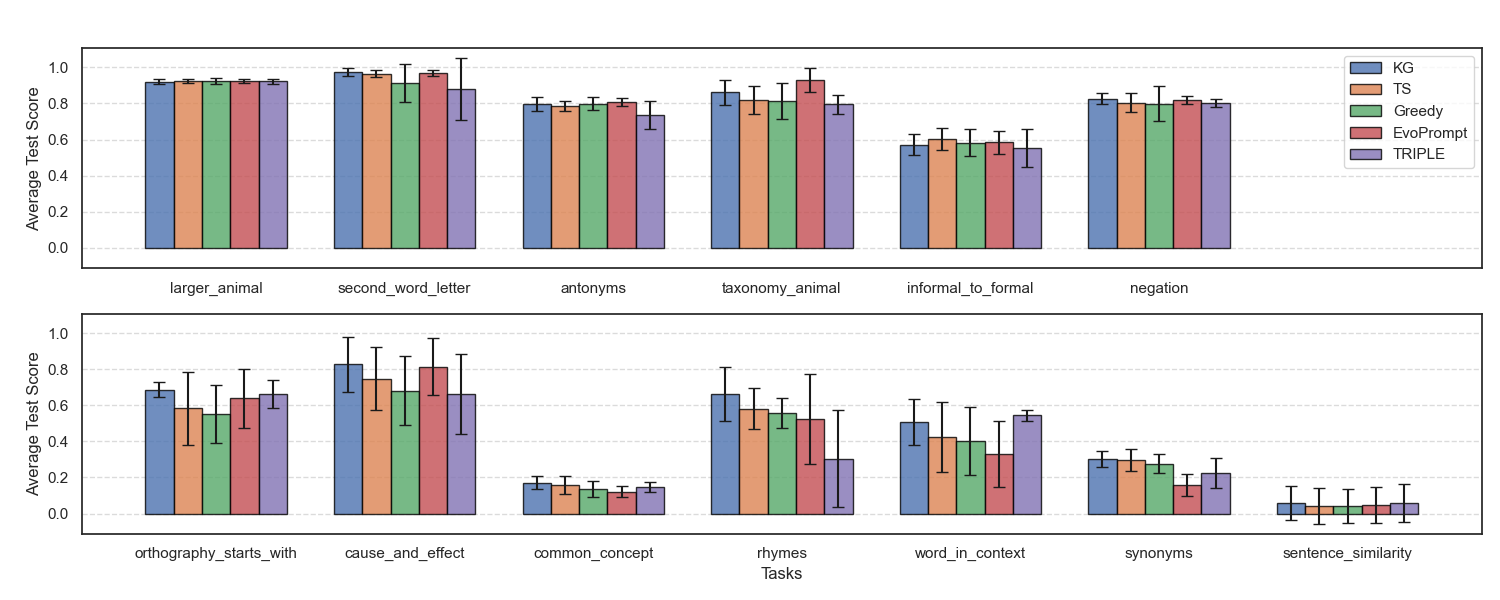}
    \caption{Test performance on 13 tasks for different methods. The height of each bar represents the average test score and the error bar represents the standard deviation across 20 replications with different random seeds}
    \label{fig:mainresult}
\end{figure*}

We focus on the 13 challenging tasks where the validation score are below 80\% with relatively large variance using the default meta prompt template in \cite{honovich2022instruction}. Table~\ref{tab:mainresult} presents the average test performance across 13 tasks for SOPL and the benchmark approaches. For each task, we illustrate the mean and standard deviation of the test score across 20 replications in Figure~\ref{fig:mainresult}. The results indicate that SOPL-KG outperforms the benchmarks. It exhibits the highest average test score of 0.6281 among all methods, with a 6.47\% improvement in average test score and a 17.92\% average improvement per task relative to EvoPrompt. For each task, we rank the five methods from the highest to the lowest test score, and calculate the average ranking across 13 tasks. The SOPL-KG achieves the highest average ranking of 1.85, and the SOPL-TS has the second best ranking of 2.69, outperforming both EvoPrompt and TRIPLE. We compute the standard deviation of the test scores across 20 replications for each task, and report the average across 13 tasks in Table~\ref{tab:mainresult}. The SOPL-KG exhibits the lowest standard deviation, demonstrating its robustness to variations in random seeds.

The findings highlight the effectiveness of our proposed framework with feature-based prompt representations and the KG prompt selection policy in comparison to other benchmarks. Although the space of all feature combinations is prohibitively large for exhaustive exploration, the SOPL-KG is capable of discovering high-quality prompts within given prompt evaluations. 

\begin{table*}
\centering
\begin{tabular}{l c c c c c}
\toprule
\textbf{Metric} & \textbf{SOPL-KG} & \textbf{EvoPrompt} & \textbf{TRIPLE} & \textbf{SOPL-TS} & \textbf{SOPL-Greedy} \\ 
\midrule
Test score & \textbf{0.6281} & 0.5900 & 0.5609 & 0.5948 & 0.5750 \\
Standard deviation & \textbf{0.0668} & 0.0881 & 0.0966 & 0.0880 & 0.0959 \\ 
Improvement of SOPL-KG & 0.00\% & 6.47\% & 11.99\% & 5.60\% & 9.23\% \\ 
Improvement of SOPL-KG per task & 0.00\% & 17.92\% & 17.19\% & 9.13\% & 14.35\% \\ 
Ranking & \textbf{1.85} & 2.92 & 3.77 & 2.69 & 3.69 \\ \bottomrule
\end{tabular}
\caption{Average test performance across 13 tasks for different methods}
\label{tab:mainresult}
\end{table*}

\subsection{Sensitivity to the Number of Iterations}
We consider more challenging scenarios with fewer iterations of $N=20$ and $N=10$. Table~\ref{tab:budget_result} presents the average test performance across 13 tasks. The SOPL-KG outperforms all other methods within fewer iterations, with a slightly lower average test score of $0.6174$ when $N=20$ compared to $N=30$. The SOPL-KG also has the lowest standard deviation when $N=20$, while the SOPL-TS is slightly more robust to variations in random seeds when $N=10$.

\begin{table}
\centering
\begin{tabular}{l c c c c }
\toprule
\multirow{2}{*}{\textbf{Method}} 
& \multicolumn{2}{c}{$N=20$} 
& \multicolumn{2}{c}{$N=10$} \\ 
& \textbf{Mean} & \textbf{STD} 
& \textbf{Mean} & \textbf{STD} \\ \midrule
EvoPrompt & 0.5776 & 0.0956 & 0.5625 & 0.0920 \\
TRIPLE & 0.5561 & 0.0996 & 0.5333 & 0.1087 \\
SOPL-KG & \textbf{0.6174} & \textbf{0.0771} & \textbf{0.5800} & 0.0935 \\
SOPL-TS & 0.5926 & 0.0845 & 0.5696 & \textbf{0.0893} \\
SOPL-Greedy & 0.5757 & 0.0941 & 0.5490 & 0.1012 \\ \bottomrule
\end{tabular}
\caption{Average test score after fewer iterations}
\label{tab:budget_result}
\end{table}

We implement an early stopping mechanism in our framework to further reduce the cost of prompt evaluation. We terminate the process early if the best validation score does not improve for $\tau$ consecutive steps, or when the maximum number $N$ of steps is reached. We conduct experiments with $\tau = 5$ and $\tau = 10$ for 20 replications and report the results in Table~\ref{tab:early_stopping_result}. Within 17 realized iterations, all three prompt selection policies yield close to but slightly worse test score compared to $N=30$. TS has a small advantage in the average number of steps used, but KG dominates the test score relative to baseline policies. It shows that the KG policy has the potential of reducing the number of evaluations required without significantly compromising performance.
\begin{table}
\centering
\begin{tabular}{lc c c c}
\toprule
\multirow{2}{*}{\textbf{Policy}} 
& \multicolumn{2}{c}{$\tau=10$} 
& \multicolumn{2}{c}{$\tau=5$} \\ 
& \textbf{Test score} & \textbf{Steps} 
& \textbf{Test score} & \textbf{Steps}  \\ \midrule
SOPL-KG & \textbf{0.6060} & 16.88 & \textbf{0.5711} & 8.45 \\ 
SOPL-TS & 0.5813 & \textbf{16.10} & 0.5488 & \textbf{8.20} \\ 
SOPL-Greedy & 0.5653 & 16.60 & 0.5389 & 8.37 \\
\bottomrule
\end{tabular}
\caption{Average number of realized iterations and average test score for different policies with early stopping}
\label{tab:early_stopping_result}
\end{table}

\subsection{Sensitivity to the Prompt Selection Policy}
\begin{figure}
    \centering
    \includegraphics[width=0.9\linewidth]{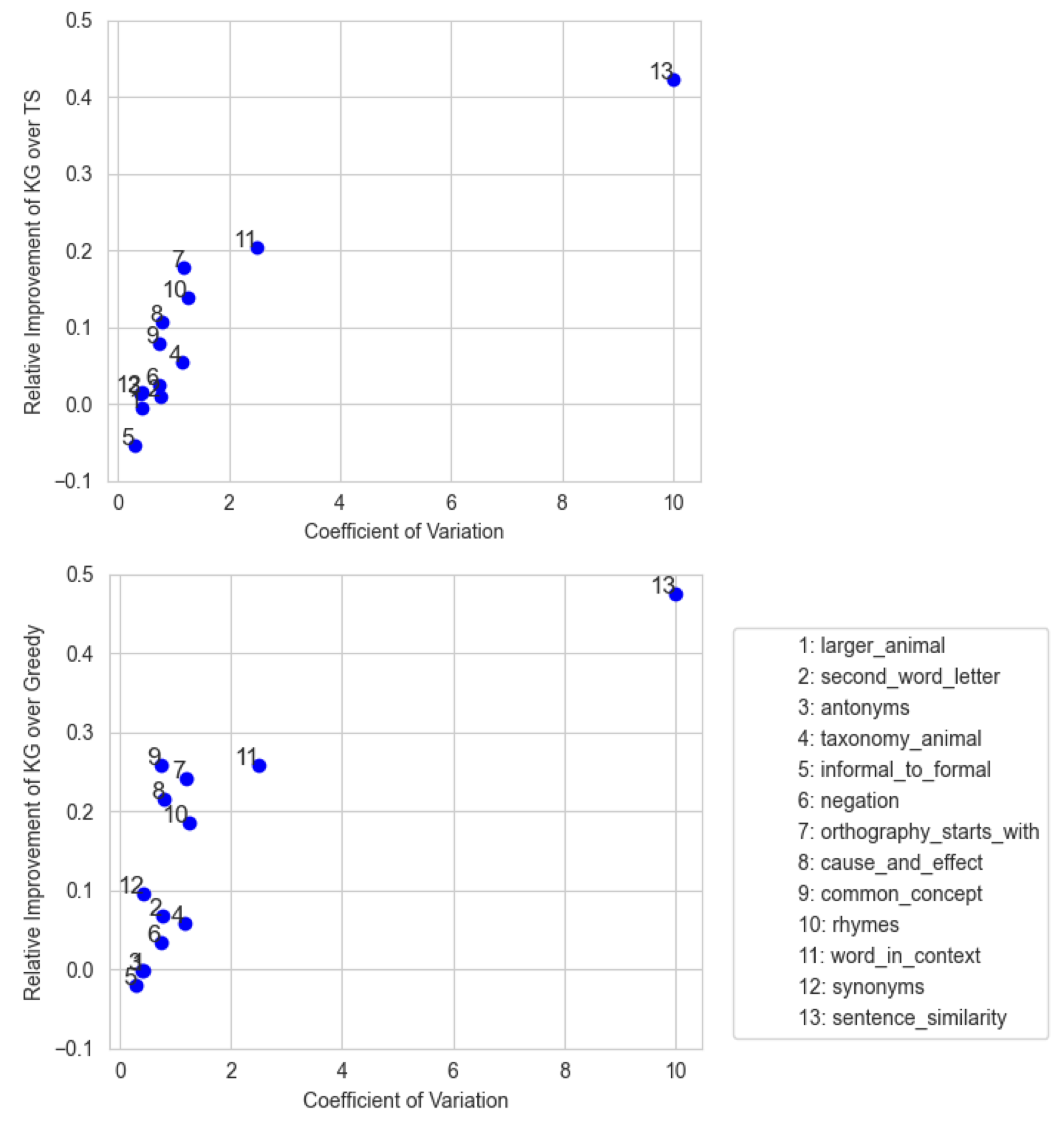}
     \caption{Upper plot: Improvement over SOPL-TS versus the coefficient of variation. Lower plot: Improvement over SOPL-Greedy versus the coefficient of variation} \label{fig:kg-vs-cv}
\end{figure}

As Figure~\ref{fig:mainresult} illustrates, the advantage of the KG policy is more evident for some tasks while more subtle for others. For each task, we randomly sample 100 feature vectors and evaluate their validation scores. We compute the mean $\mu_{100}$ and the standard deviation $\sigma_{100}$ of the 100 scores, and calculate the coefficient of variation by $CV_{100} := \sigma_{100}/\mu_{100}$. This metric aims to represent the uncertainty in the LLM response performance as the prompt feature values vary. Figure~\ref{fig:kg-vs-cv} depicts the relationship between the relative improvement of the KG prompt representation selection policy over the TS and Greedy policies in the average test score, and the coefficient of variation $CV_{100}$ for different tasks. The correlation coefficient between $CV_{100}$ and the relative improvement of KG over TS and Greedy across 13 tasks are $\rho_1 = 0.8901$ and $\rho_2 = 0.7789$. The positive correlations suggest that for tasks where the LLM response is highly sensitive to prompt features, the KG policy can prominently outperform TS and Greedy in the SOPL framework.

We also observed that for easier tasks when the LLM response is relatively insensitive to the prompts and the performance function is relatively flat with respect to prompt features, full exploitation policies such as Greedy are adequate to identify an effective prompt. In particular, we observe that Greedy outperforms KG by a very small margin for two easier tasks, \texttt{antonyms} and \texttt{informal\_to\_formal}. However, for challenging tasks with high uncertainty, KG consistently yields over a 10\% improvement relative to both baselines when other parts of the framework remain the same. Our analysis reveals the importance of the choice of the policy depending on the problem context.

\subsection{Sensitivity to Feature Selection}
\begin{figure}
    \centering
    \includegraphics[width=0.75\linewidth]{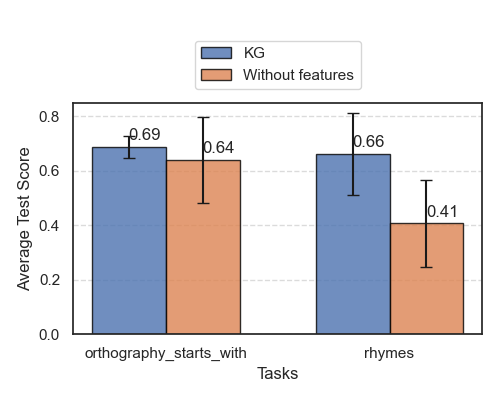}
    \caption{Comparison of the average test score between our proposed method using all five features and the method using only one feature for the demonstrative examples}
    \label{fig:feature-imp}
\end{figure}

We assess the effectiveness of enriching the prompt representation by multiple features on the two tasks with the largest improvement of SOPL-KG compared to the second best performance, i.e., \texttt{orthography\_starts\_with} and \texttt{rhymes}. We use the default meta prompt from \cite{honovich2022instruction}, which only includes the feature for the required demonstration examples, and use SOPL-KG to select from 20 predefined configurations of demonstration examples. We allow $N=30$ evaluations and repeat the experiments for 20 replications. Figure~\ref{fig:feature-imp} shows that the performance deteriorates as the features that enhance the meta prompt are excluded. While searching in a single dimension is easier than optimizing multiple features simultaneously, it results in finding suboptimal prompts. The result suggests that enriching the meta prompt with different features effectively expands the search space and leads to improved performance of the generated prompts.
 
\section{Conclusion and Future Work}\label{sec:conclusion}
This paper introduces SOPL, a sequential optimal prompt learning framework for automated prompt engineering focused on efficient prompt learning in practical scenarios where exhaustive evaluation is costly or impossible. Specifically, we develop a feature-based approach to model prompts, enabling a constraint-based and expansive prompt search space. The forward-looking KG policy with correlated beliefs facilitates efficient and scalable prompt learning. We demonstrate that our proposed method achieves superior performance on instruction induction tasks with only 30 or fewer opportunities of prompt evaluation. We find that the KG policy yields substantial performance gains compared to baseline policies, especially for challenging tasks with high uncertainty. Moreover, our framework allows for future investigation of continuous representations of prompts by embedding vectors. Our work shows a promising direction of leveraging optimal learning methods for efficient prompt learning, paving the way for future research on scalable prompt engineering.

\section*{Acknowledgment}\label{sec:ack}
This research was supported in part through the computational resources and staff contributions provided for the Quest high performance computing facility at Northwestern University which is jointly supported by the Office of the Provost, the Office for Research, and Northwestern University Information Technology.

\clearpage

\bibliographystyle{plain}
\bibliography{reference}

\end{document}